\documentclass[11pt,letterpaper]{article}
\usepackage{naaclhlt2016}
\usepackage{times}
\usepackage{latexsym}
\usepackage{url}
\usepackage{multirow}
\usepackage{amssymb}
\usepackage{amsthm}
\usepackage{amsmath}
\usepackage{verbatim}
\usepackage{tikz-qtree}
\usepackage{tikz}
\usetikzlibrary{shapes.geometric}
\usetikzlibrary{shapes.arrows}
\usetikzlibrary{patterns}
\usetikzlibrary{chains}
\usetikzlibrary{calc}
\usepackage{tipa}
\usepackage[utf8]{inputenc}
\usepackage[T2A]{fontenc}
\usepackage[russian, english]{babel}
\usepackage{xcolor}
\usepackage{graphicx}
\usepackage{scalerel}
\usepackage{breqn}
\usepackage{url}

\naaclfinalcopy 
\def\naaclpaperid{370} 

\newcommand{\ignore}[1]{}
\newcommand{\Sref}[1]{\S\ref{#1}}
\newcommand{\fref}[1]{figure~\ref{#1}}
\newcommand{\Fref}[1]{Figure~\ref{#1}}
\newcommand{\tref}[1]{table~\ref{#1}}

\usepackage[T1]{fontenc}

\newcommand{\mfar}[1]{\textcolor{red}{\bf\small [#1 --MF]}}

\title{Polyglot Neural Language Models: \\A Case Study in Cross-Lingual Phonetic Representation Learning}

 \author{Yulia Tsvetkov \quad Sunayana Sitaram \quad Manaal Faruqui \quad Guillaume Lample \\
        \textbf{Patrick Littell \quad David Mortensen \quad Alan W Black \quad Lori Levin \quad Chris Dyer}\\
         Language Technologies Institute \\ Carnegie Mellon University \\ Pittsburgh, PA, 15213, USA\\
          { \{ytsvetko,ssitaram,mfaruqui,glample,plittell,dmortens,awb,lsl,cdyer\}@cs.cmu.edu}\\}

\date{}

\begin{document}

\maketitle

\begin{abstract}
We introduce polyglot language models, recurrent neural network models trained to predict symbol sequences in many different languages using shared representations of symbols and conditioning on typological information about the language to be predicted.
We apply these to the problem of modeling phone sequences---a domain in which universal symbol inventories and cross-linguistically shared feature representations are a natural fit.
Intrinsic evaluation on held-out perplexity, qualitative analysis of the learned representations, and extrinsic evaluation in two downstream applications that make use of phonetic features show (i) that polyglot models better generalize to held-out data than comparable monolingual models and (ii) that polyglot phonetic feature representations are of higher quality than those learned monolingually.
\end{abstract}

\section{Introduction}
\ignore{Statistical language models (LMs) are a core component of machine translation,
speech recognition, information retrieval, and other language processing tasks,
that estimates semantic and morphosyntactic fluency of a sequence in a language.
Traditional LMs are computed as $n$-gram counts with 
smoothing \cite{kneser1995improved,chen1996empirical}.  %
More recently, neural probabilistic language models (NLMs) have been shown to
outperform the count-based models, due to their ability to share learn and share representations across many different word types. 
Most prominent NLM architectures  include
feed-forward networks \cite{bengio2003lm,schwenk2007lm},
log-bilinear models \cite{mnih2007lm},
and recurrent neural networks \cite{mikolov2010rnnlm,mikolov2011extensions,sundermeyer2012lstm}.}

Nearly all existing language model (LM) architectures are designed to model one language at a time.
This is unsurprising considering the historical importance of count-based models in which every surface form of a word
is a separately modeled entity (English \textit{cat} and Spanish \textit{gato} would not
likely benefit from sharing counts). However, recent models that use distributed representations---in particular models that share representations across languages \cite[\emph{inter alia}]{Hermann:2014:ACLphil,faruqui:2014,huang-EtAl:2015:EMNLP,lu:2015}---suggest universal models applicable to multiple languages are a possibility.
This paper takes a step in this direction.

We introduce \textbf{polyglot language models}: neural network language models that are trained on
and applied to any number of languages. Our goals with these models are the following. First, to facilitate data and parameter sharing,
providing more training resources to languages, which is especially valuable in low-resource settings.
Second, models trained on diverse languages with diverse linguistic properties
will better be able to learn naturalistic representations that are less likely to 
``overfit'' to a single linguistic outlier. 
Finally, polyglot models offer convenience in a multilingual world: a single model replaces dozens of different models.

Exploration of polyglot language models at the sentence level---the traditional domain of language modeling---requires dealing with a massive event space (i.e., the union of words across many languages).
To work in a more tractable domain, we evaluate our model on
phone-based language modeling, the modeling sequences of \emph{sounds}, rather than words. 
We choose this domain since a common assumption of many theories of phonology is 
that all spoken languages construct words from a finite inventory of phonetic symbols 
(represented conveniently as the elements of the the International Phonetic Alphabet; IPA) 
which are distinguished by language-universal features (e.g., place and manner of articulation, voicing status, etc.).
Although our focus is on sound sequences, our solution can be ported to the semantic/syntactic
problem as resulting from adaptation to constraints on semantic/syntactic  structure.

This paper makes two primary contributions: in modeling and in applications.
In \Sref{model}, we introduce a novel polyglot neural language model (NLM) architecture.
Despite being trained on multiple languages, the multilingual model is more effective (9.5\% lower
perplexity) than individual models, and substantially more effective than naive baselines (over 25\% lower perplexity). Our most effective polyglot architecture conditions not only on the identity of the language being predicted in each sequence, but also on a vector representation of its phono-typological properties.
In addition to learning representations of phones as part of the polyglot language modeling objective, the model incorporates features about linguistic typology to improve generalization performance  (\Sref{typology}).
Our second primary contribution is to show that downstream applications are improved by using polyglot-learned phone representations. We focus on two tasks: predicting adapted word forms in models of cross-lingual lexical borrowing and speech synthesis (\Sref{downstream}).
Our experimental results (\S\ref{sec:exp}) show that in borrowing, we improve over the current state-of-the-art, and in speech synthesis, our features are more effective than manually-designed phonetic features.
Finally, we analyze the phonological content of learned representations, 
finding that our polyglot models discover standard phonological categories such as length and nasalization, and that these are grouped correctly across languages with different phonetic inventories and contrastive features.

\section{Model}
\label{model}
In this section, we first describe in \Sref{rnnlm} the underlying
framework of our model---RNNLM---a standard recurrent neural network based language
model \cite{mikolov2010rnnlm,sundermeyer2012lstm}. Then, in \Sref{polylm}, we define a 
Polyglot LM---a modification of RNNLM to incorporate language information, both learned
and hand-crafted. 


\paragraph{Problem definition.}
In the phonological LM, \textit{phones} (sounds) are the basic units. Mapping from
words to phones is defined in pronunciation dictionaries.
For example, ``cats'' \textipa{[k\ae ts]} is a sequence of four phones.
Given a prefix of phones $\phi_{1}, \phi_{2},\ldots, \phi_{t-1}$, 
the task of the LM is to estimate the conditional probability of the next phone $p(\phi_{t} \mid \phi_{1}, \phi_{2},\ldots,\phi_{t-1})$.

\subsection{RNNLM}
\label{rnnlm}
In NLMs, a vocabulary $V$ (here, a set of phones composing all word types in the language)
is represented as a matrix of parameters $\mathbf{X} \in \mathbb{R}^{d \times |V|}$,
with $|V|$ phone types represented as $d$-dimensional vectors.  $\mathbf{X}$ is
often denoted as lookup table.
Phones in the input sequence are first converted to phone vectors, where $\phi_i$ is represented by $\mathbf{x}_i$ by multiplying the phone indicator (one-hot vector of length $|V|$) and the lookup table.

At each time step $t$, most recent phone prefix vector\footnote{ We are reading at each time step 
the most recent $n$-gram context rather than---as is more common in RNNLMs---a single phone context. 
Empirically, this works better for phone sequences, and we hypothesize that this lets the learner 
rely on direct connections for local phenomena (which are abundant in phonology) and 
minimally use the recurrent state to model longer-range effects.} 
$\mathbf{x}_{t}$ and hidden state $\mathbf{h}_{t-1}$ are transformed to compute a new hidden representation:
\begin{align*}
\mathbf{h}_{t} =f(\mathbf{x}_{t}, \mathbf{h}_{t-1}) , 
\end{align*}
where $f$ is a non-linear transformation. 
In the original RNNLMs \cite{mikolov2010rnnlm}, the transformation is such that: 
\begin{align*}
\mathbf{h}_{t} =\mathrm{tanh}(\mathbf{W}_{h_x}\mathbf{x}_{t} + \mathbf{W}_{h_h}\mathbf{h}_{t-1}+\mathbf{b}_{h}) .
\end{align*}
To overcome the notorious problem in recurrent neural networks of vanishing
gradients \cite{bengio1994learning}, following \newcite{sundermeyer2012lstm}, 
in recurrent layer we use long short-term
memory (LSTM) units \cite{hochreiter:1997}:\footnote{For brevity, we omit the
equations describing the LSTM cells; they can be found in \cite[eq. 7--11]{graves:2013}.}
\begin{align*}
\mathbf{h}_{t} =\mathrm{LSTM}(\mathbf{x}_{t}, \mathbf{h}_{t-1}) . 
\end{align*}

Given the hidden sequence $\mathbf{h}_{t}$, the output sequence is then computed as follows:
\begin{align*}
p(\phi_t = i \mid \phi_{1},&\ldots,\phi_{t-1}) = \\
&\mathrm{softmax}(\mathbf{W}_{out}\mathbf{h}_{t} + \mathbf{b}_{out})_i,
\end{align*}
where $\mathrm{softmax}(x_i) = \frac{e^{x_i}}{\sum_je^{{x}_j}}$ ensures a valid
probability distribution over output phones.

\subsection{Polyglot LM}
\label{polylm}
\begin{figure}[!ht]
  \centering
  \includegraphics[width=1\columnwidth]{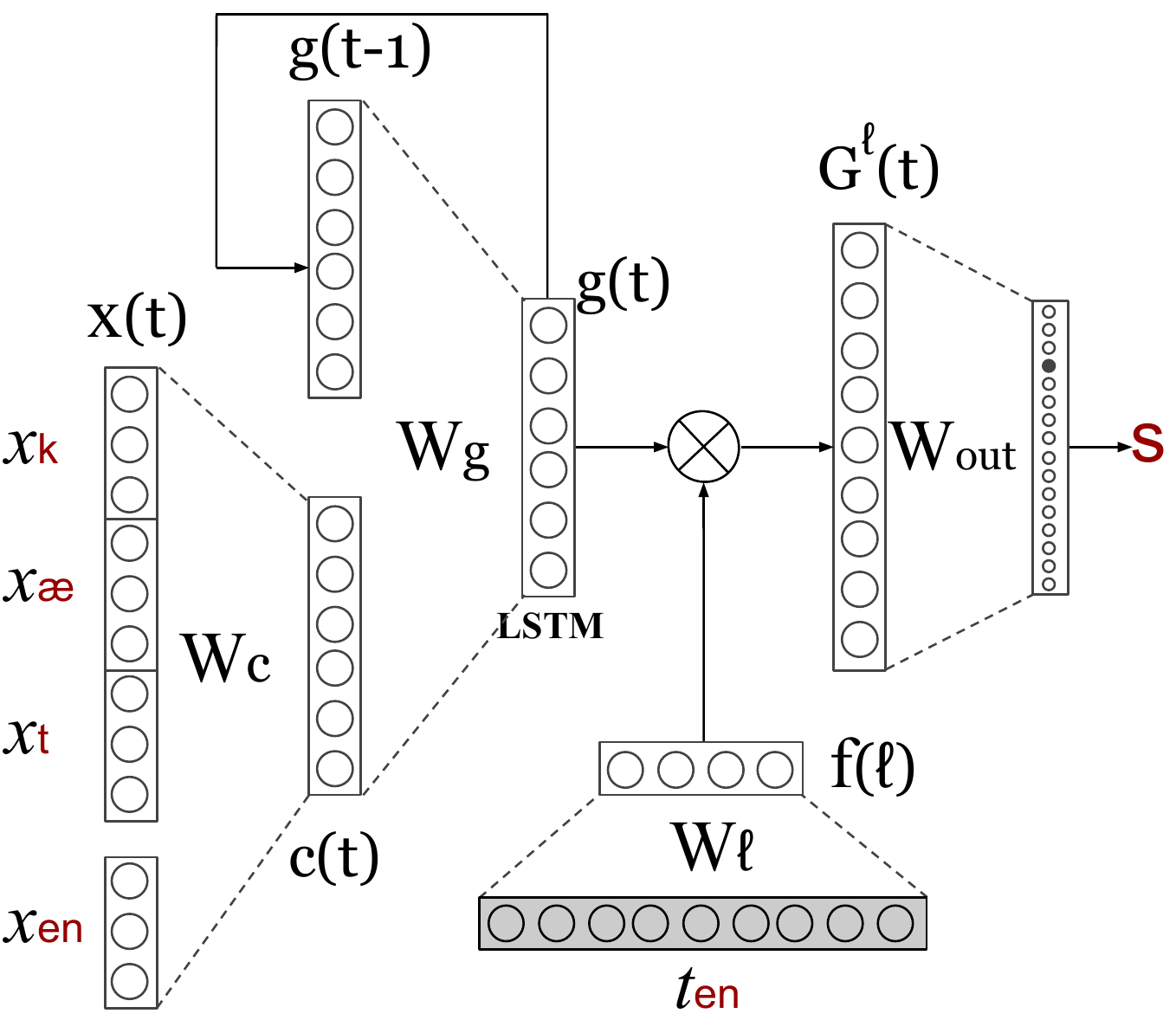}
  \caption{Architecture of the Polyglot LM.}
  \label{fig:architecture}
\end{figure}

We now describe our modifications to RNNLM to account for multilinguality.
The architecture is depicted in \fref{fig:architecture}. Our task is to  estimate the conditional probability of the next phone given the preceding phones and the language ($\ell$):  $p(\phi_{t} \mid \phi_{1}, \ldots, \phi_{t-1}, \ell)$.

In a multilingual NLM, we define a vocabulary  $V^*$ to be the union of vocabularies
of all training languages, assuming that all language vocabularies are mapped to a
shared representation (here, IPA). 
In addition, we maintain $V_{\ell}$ with a special
symbol for each language (e.g., $\phi_{\textit{english}}$, $\phi_{arabic}$). 
Language symbol vectors are parameters in the new lookup table $\mathbf{X}_{\ell} \in \mathbb{R}^{d \times |\# langs|}$
(e.g., $\mathbf{x}_{\textit{english}}$, $\mathbf{x}_{\textit{arabic}}$). 
The inputs to the Polyglot LM are the phone vectors $\mathbf{x}_{t}$, 
the language character vector $\mathbf{x}_{\ell}$, and the typological feature vector constructed externally $\mathbf{t}_{\ell}$. The typological feature vector will be discussed in the following section.

The input layer is passed to the hidden local-context layer:
\begin{align*}
\mathbf{c}_{t} =\mathbf{W}_{c_x}\mathbf{x}_{t} + \mathbf{W}_{c_{lang}}\mathbf{x}_{lang}+\mathbf{b}_{c} .
\end{align*}
The local-context vector is then passed to the hidden LSTM
global-context layer, similarly to the previously described RNNLM:
\begin{align*}
\mathbf{g}_{t} =\mathrm{LSTM}(\mathbf{c}_{t}, \mathbf{g}_{t-1}) . 
\end{align*}

In the next step, the global-context vector $\mathbf{g}_{t}$ is
``factored'' by the typology of the training language, to integrate
manually-defined language features. To obtain this, we first project
the (potentially high-dimensional) $\mathbf{t}_{\ell}$ into a
low-dimensional vector, and apply non-linearity.
Then, we multiply the $\mathbf{g}_{t}$ and the projected language layer,
to obtain a global-context-language matrix:\ignore{\footnote{Gating the context
representation by auxiliary features is similar in spirit to integrating
modality in the multimodal neural language models introduced by
\newcite{kiros2013multimodal}.
Our architecture, however, is substantially different;
in \Sref{related}, we clarify the difference between the two models.}}

\begin{align*}
\mathbf{f}_{\ell} &=\mathrm{tanh}(\mathbf{W}_{\ell}\mathbf{t}_{\ell} + \mathbf{b}_{\ell}) ,\\
\mathbf{G}^{\ell}_{t} &= \mathbf{g}_{t}\otimes \mathbf{f}_{\ell}^{\top} .
\end{align*}

Finally, we vectorize the resulting matrix into a column vector and compute the output sequence as follows:
\begin{align*}
p(\phi_t = i \mid \phi_1,& \ldots, \phi_{t-1},\ell) =\\
&\mathrm{softmax}(\mathbf{W}_{\textit{out}}\mathrm{vec}(\mathbf{G}^{\ell}_{t}) + \mathbf{b}_{\textit{out}})_i .
\end{align*}

\paragraph{Model training.}
Parameters of the models are the lookup tables $\mathbf{X}$ and $\mathbf{X}_{\ell}$, weight matrices
$\mathbf{W}_{i}$, and bias vectors $\mathbf{b}_{i}$.
Parameter optimization is performed using stochastic updates 
to minimize the categorical cross-entropy loss (which is
equivalent to minimizing perplexity and maximizing likelihood):
$H(\phi, \hat{\rule{0ex}{1.4ex}\smash{\phi}}) = -\Sigma_{i} \hat{\rule{0ex}{1.4ex}\smash{\phi}}_{i}\ \mathrm{log}\  \phi_{i}$,
where $\phi$ is predicted and $\hat{\rule{0ex}{1.4ex}\smash{\phi}}$ is the gold label.

\section{Typological features}
\label{typology}
Typological information is fed to the model via vectors of 190
binary typological features, all of which are phonological (related to sound
structure) in their nature.
These feature vectors are derived from
data from the WALS \cite{wals2013}, PHOIBLE \cite{phoible2014}, and 
Ethnologue \cite{ethnologue2015}  
typological databases via extensive post-processing and
analysis.\footnote{This data resource, which provides standardized phono-typological information for 2,273 languages, is available at \url{https://github.com/dmort27/uriel-phonology/ tarball/0.1}. It is a subset of the URIEL database, a comprehensive database of typological features encoding syntactic and morphological (as well as phonological) properties of languages. It is available at \url{http://cs.cmu.edu/~dmortens/uriel.html}.} 
The features primarily concern properties of sound
inventories (i.e., the set of phones or phonemes occurring in a language)
and are mostly of one of four types:
\begin{enumerate}\itemsep1pt \parskip0pt \parsep0pt
\item \textbf{Single segment represented in an inventory}; e.g., does
  language $\ell$'s sound inventory include /\textipa{g}/, a voiced velar
  stop?
\item \textbf{Class of segments represented in an
    inventory}; e.g., does language $\ell$'s sound inventory include voiced
  fricatives like /\textipa{z}/ and /\textipa{v}/?
\item \textbf{Minimal contrast represented in an inventory}; e.g.,
  does language $\ell$'s sound inventory include two sounds that differ only
  in voicing, such as /\textipa{t}/ and /\textipa{d}/?
\item \textbf{Number of sounds representative of a class that are present in
    an inventory}; e.g., does language $\ell$'s sound inventory include
  exactly five vowels?
\end{enumerate}

\ignore{Once the programmatic infrastructure was place, these features could
be deduced for any language for which the sound inventory was
available in computer-readable form; however, developing this
infrastructure was not trivial. While the code that processes WALS and
PHOIBLE to produce these features is computationally straightforward,}
The motivation and criteria for coding each individual feature
required extensive linguistic knowledge and analysis. Consider the
case of tense vowels like /\textipa{i}/ and /\textipa{u}/ in ``beet''
and ``boot'' in contrast with lax vowels like /\textipa{I}/ and
/\textipa{U}/ in ``bit'' and ``book.'' Only through linguistic
analysis does it become evident that (1)~all languages have tense
vowels---a feature based on the presence of tense vowels is
uninformative and that (2)~a significant minority of languages make a
distinction between tense and lax vowels---a feature based on whether
languages display a minimal difference of this kind would be more
useful.

\section{Applications of Phonetic Vectors}
\label{downstream}
Learned continuous word representations---word vectors---are an important by-product
of neural LMs, and these are used as features in numerous NLP applications, including
chunking \cite{turian:2010},
part-of-speech tagging \cite{ling2015finding},
dependency parsing \cite{lazaridou:2013,bansal:2014,dyer2015transition,watanabe2015transition},
named entity recognition \cite{guo2014revisiting},
and sentiment analysis \cite{socher:2013,wang2015predicting}.
We evaluate phone vectors learned by Polyglot LMs in
two downstream applications that rely on phonology:
modeling lexical borrowing (\S\ref{borrowing}) and speech synthesis (\S\ref{voice}).

\subsection{Lexical borrowing}
\label{borrowing}
Lexical borrowing is the adoption of words from another language,
that inevitably happens when speakers of different languages communicate
for a long period of time \cite{thomason2001language}.
Borrowed words---also called \textit{loanwords}---constitute 
10--70\% of most language lexicons \cite{haspelmath2009lexical};
these are content words of foreign origin that are adapted in
the language and are not perceived as foreign by language speakers.
Computational modeling of cross-lingual transformations of loanwords is effective for inferring lexical correspondences across languages with limited parallel data, benefiting applications such as machine translation \cite{borrowing-acl,borrowing-jair}.

In the process of their nativization in a foreign language,
loanwords undergo primarily \textbf{phonological adaptation}, namely
insertion/deletion/substitution of phones to adapt to the
phonotactic constraints of the recipient language. 
If a foreign phone is not present in the recipient language,
it is usually replaced with its closest native equivalent---we thus hypothesize that cross-lingual phonological features learned by the
Polyglot LM can be useful in models of borrowing
to quantify cross-lingual similarities of sounds.

\begin{figure}[!ht]
  \centering
  \includegraphics[width=1\columnwidth]{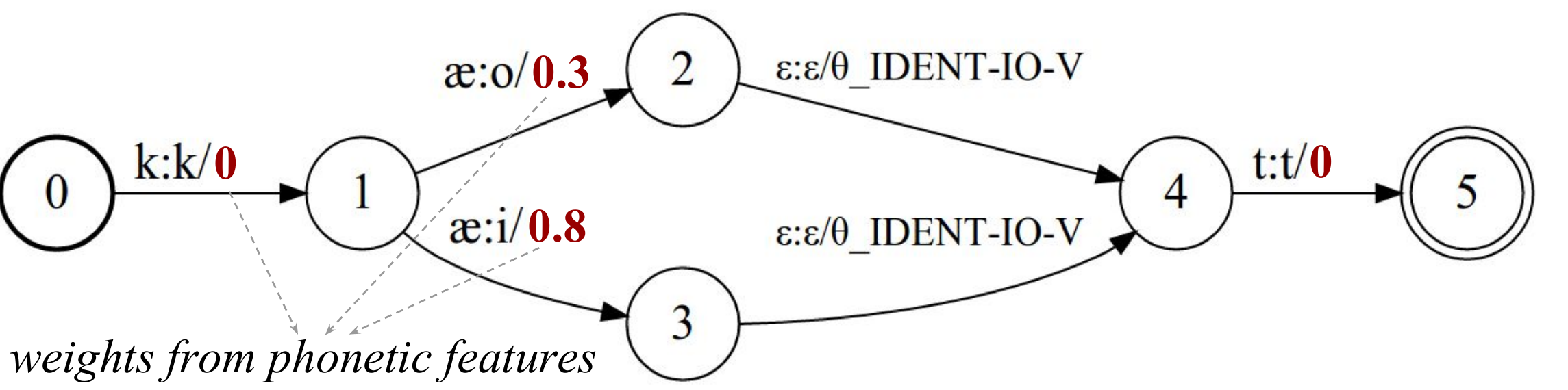}
  \caption{Distances between phone vectors learned by the Polyglot LM are integrated
  as substitution weights in the lexical borrowing transducers. 
  An English word \textit{cat} [\textipa{k\ae t}] is adapted to its Russian
  counterpart \foreignlanguage{russian}{\textit{кот}} [kot].
  The transducer has also an erroneous path to \foreignlanguage{russian}{\textit{кит}} [kit] `whale'.
  In the original system, both paths are weighted with the same feature \textsc{ident-io-v},
  firing on vowel substitution. Our modification allows the borrowing model to identify more plausible
  paths by weighting substitution operations.}
  \label{fig:in_borrowing}
\end{figure}

To test this hypothesis, we augment the hand-engineered models proposed by \newcite{borrowing-jair}
with features from phone vectors learned by our model.
Inputs to the borrowing framework are loanwords (in Swahili, Romanian, Maltese),
and outputs are their corresponding ``donor'' words in the donor language (Arabic, French, Italian, resp.).
The framework is implemented as a cascade of finite-state transducers with
insertion/deletion/substitution operations on sounds, weighted by
high-level conceptual linguistic constraints that are learned in a supervised manner.
Given a loanword, the system produces a candidate donor word
with lower ranked violations than other candidates, using the shortest path algorithm.
In the original borrowing model, insertion/deletion/substitution operations are unweighted.
In this work, we integrate transition weights in the phone substitution transducers, 
which are cosine distances between phone vectors learned by our model.
Our intuition is that similar sounds appear in similar contexts,
even if they are not present in the same language (e.g., /\textipa{s\super Q}/
in Arabic is adapted to /\textipa{s}/ in Swahili).
Thus, if our model effectively captures cross-lingual signals,
similar sounds should have smaller distances in the vector space, which
can improve the shortest path results.
\Fref{fig:in_borrowing} illustrates our modifications to the original framework.

\subsection{Speech synthesis}
\label{voice}
Speech synthesis is the process of converting text into speech. It has various applications, such as screen readers for the visually impaired and hands-free voice based systems. Text-to-speech (TTS) systems are also used as part of speech-to-speech translation systems and spoken dialog systems, such as personal digital assistants.
Natural and intelligible TTS systems exist for a number of languages in the world today. However, building TTS systems remains prohibitive for many languages due to the lack of linguistic resources and data.

\ignore{\mfar{can be removed}
Many of these low resource languages have a high user population, or a large number of non-literate speakers for which developing speech-based systems may be beneficial.}

The language-specific resources that are traditionally used for building TTS systems in a new language are:
(1) audio recordings with transcripts; (2) pronunciation lexicon or letter to sound rules; and (3) a phone set definition.
Standard TTS systems today use phone sets designed by experts. Typically, these phone sets also contain phonetic features for each phoneme, which are used as features in models of the spectrum and prosody. The phonetic features available in standard TTS systems are multidimensional vectors indicating various properties of each phoneme, such as whether it is a vowel or consonant, vowel length and height, place of articulation of a consonant, etc. Constructing these features by hand can be labor intensive, and coming up with such features automatically may be useful in low-resource scenarios.


In this work, we replace manually engineered phonetic features with phone vectors, which are then used by classification and regression trees for modeling the spectrum. Each phoneme in our phone set is assigned an automatically constructed phone vector, and each member of the phone vector is treated as a phoneme-level feature which is used in place of the manually engineered phonetic features.
While prior work has explored TTS augmented with acoustic features \cite{watts2015sentence},
to the best of our knowledge, we are the first to replace manually
engineered phonetic features in TTS systems with automatically constructed phone vectors.


\section{Experiments}\label{sec:exp}
Our experimental evaluation of our proposed polyglot models consists of two parts: (i) an intrinsic evaluation where phone sequences are modeled with independent models and (ii) an extrinsic evaluation of the learned phonetic representations. Before discussing these results, we provide details of the data resources we used.
\label{experiments}
\subsection{Resources and experimental setup}
\paragraph{Resources.}
We experiment with the following languages:
Arabic (\textsc{ar}), French (\textsc{fr}), Hindi (\textsc{hi}),
Italian (\textsc{it}), Maltese (\textsc{mt}), Romanian (\textsc{ro}),
Swahili (\textsc{sw}), Tamil (\textsc{ta}), and Telugu (\textsc{te}).
In our language modeling experiments, two main sources of data are pronunciation
dictionaries and typological features described in \Sref{typology}.
The dictionaries for \textsc{ar}, \textsc{fr}, \textsc{hi}, \textsc{ta}, and \textsc{te}
are taken from in-house speech recognition/synthesis systems.
For remaining languages, the dictionaries are automatically constructed
using the Omniglot grapheme-to-IPA conversion rules.\footnote{\url{http://omniglot.com/writing/}}

We use two types of pronunciation dictionaries:
(1) \textsc{ar}, \textsc{fr}, \textsc{hi},
\textsc{it}, \textsc{mt}, \textsc{ro}, and \textsc{sw}
dictionaries used in experiments with lexical borrowing; and
(2) \textsc{en}, \textsc{hi}, \textsc{ta}, and \textsc{te}
dictionaries used in experiments with speech synthesis.
The former are mapped to IPA, with the resulting phone vocabulary size---the number
of distinct phones across IPA dictionaries---of 127 phones.
The latter are encoded using the UniTran universal transliteration resource \cite{qian2010python},\ignore{\footnote{UniTran transliterates Unicode code points into phonemes from the XSAMPA phone set. Our implementation of the UniTran front end assigns a single phoneme to a Unicode character, which can be inaccurate if the language has ambiguities in letter-to-sound conversion. In case of the Indian languages, this front end does not take care of phenomena such as schwa deletion in Hindi, voicing ambiguities in Tamil etc. In case of English, it assigns a single phoneme to letters with ambiguities in pronunciation, such as 'c'. We used a list of word types from the TTS databases for Hindi, Tamil and Telugu from the IIIT-H voice databases \cite{Prahallad2012} and the RMS US English voice \cite{Kominek2004} databases to create the pronunciation dictionaries by looking up the letter-to-sound rules in UniTran front end.} }
with a vocabulary of 79 phone types.

\begin{table*}[!htb]\small
\centering
\begin{tabular}{lrrrrrrr}
 & \multicolumn{1}{c}{\textsc{ar}}
 & \multicolumn{1}{c}{\textsc{fr}}
 & \multicolumn{1}{c}{\textsc{hi}}
 & \multicolumn{1}{c}{\textsc{it}}
 & \multicolumn{1}{c}{\textsc{mt}}
 & \multicolumn{1}{c}{\textsc{ro}}
 & \multicolumn{1}{c}{\textsc{sw}}        \\\hline
train & 1,868/18,485 & 238/1,851 & 193/1,536 & 988/901 & 114/1,152 & 387/4,661 & 659/7,239 \\
dev   & 366/3,627    & 47/363    & 38/302    & 19/176  & 22/226    & 76/916    & 130/1,422 \\
test  & 208/2,057    & 27/207    & 22/173    & 11/100  & 13/128    & 43/524    & 73/806
\end{tabular}
\caption{Train/dev/test counts for IPA pronunciation dictionaries for words (phone sequences) and phone tokens, in thousands: \protect\linebreak \#thousands of sequences/\# thousands of tokens.}
\label{tbl:ipa-stats}
\end{table*}
From the (word-type) pronunciation dictionaries, we remove 
15\% of the words for development, and a further 10\% for testing; the rest of the data
is used to train the models.
In tables~\ref{tbl:ipa-stats}~and~\ref{tbl:unitran-stats}
we list---for both types of pronunciation dictionaries---train/dev/test
data statistics for words (phone sequences) and phone tokens.
We concatenate each phone sequence with beginning and end
symbols (\textit{<s>}, \textit{</s>}).

\begin{table}[]\small
\centering
\begin{tabular}{lrrrrrrr}
 & \multicolumn{1}{c}{\textsc{en}}
 & \multicolumn{1}{c}{\textsc{hi}}
 & \multicolumn{1}{c}{\textsc{ta}}
 & \multicolumn{1}{c}{\textsc{te}}\\\hline
train & 101/867 & 191/1,523 & 74/780  & 71/690  \\
dev   & 20/169  & 37/300    & 14/152  & 14/135  \\
test  & 11/97   & 21/171    & 8/87    &  8/77
\end{tabular}
\caption{Train/dev/test statistics for UniTran pronunciation dictionaries for words (phone sequences) and phone tokens, in thousands: \#thousands of sequences/\# thousands of tokens.}
\label{tbl:unitran-stats}
\end{table}

\paragraph{Hyperparameters.}
We used the following network architecture:
100-dimensional phone vectors, with hidden local-context and LSTM layers
of size 100, and hidden language layer of size 20.
All language models were trained using the left context of 3 phones (4-gram LMs). 
Across all language modeling experiments,
parameter optimization was performed on the dev set
using the Adam algorithm~\cite{kingma2014adam} with
mini-batches of size 100 to train the models for 5 epochs. 
\ignore{
All neural networks were trained on GPUs using
the Theano toolkit \cite{bergstra2010theano}.}

\subsection{Intrinsic perplexity evaluation}
\label{ppl}
Perplexity is the standard evaluation measure for language models,
which has been shown to correlate strongly with error rates in downstream applications~\cite{klakow2002testing}. 
We evaluated perplexities across several architectures, and several monolingual and multilingual setups.
We kept the same hyper-parameters across all setups, as detailed in \Sref{experiments}.
Perplexities of LMs trained on the two types of pronunciation dictionaries were evaluated separately;
\tref{tab:ppl-it} summarizes perplexities of the models trained on IPA dictionaries, and
\tref{tab:ppl-hi} summarizes perplexities of the UniTran LMs.

In columns, we compare three model architectures:
\textit{baseline} denotes the standard RNNLM architecture described in \Sref{rnnlm};
\textit{+lang} denotes the Polyglot LM architecture described in \Sref{polylm} with input language vector but without typological features and  language layer;
finally, \textit{+typology} denotes the full Polyglot LM architecture. This setup lets us separately evaluate the contribution of modified architecture and the contribution of auxiliary set of features introduced via the language layer.

Test languages are \textsc{it} in \tref{tab:ppl-it}, and \textsc{hi} in \tref{tab:ppl-hi}.
The rows correspond to different sets of training languages for the models:
\textit{monolingual} is for training and testing on the same language;
\textit{+similar} denotes training on three typologically similar languages: \textsc{it, fr, ro} in \tref{tab:ppl-it}, and \textsc{hi, ta, te} in \tref{tab:ppl-hi};
\textit{+dissimilar} denotes training on four languages, three similar and one
typologically dissimilar language, to evaluate robustness of multilingual systems
to diverse types of data. The final sets of training languages are
\textsc{it, fr, ro, hi} in \tref{tab:ppl-it}, and \textsc{hi, ta, te, en} in \tref{tab:ppl-hi}.

\ignore{
\begin{table}[ht]
\centering
\begin{tabular}{lccr}
&       \multicolumn{3}{c}{Perplexity ($\downarrow$)}     \\
\multicolumn{1}{c}{}  & baseline      & +lang         &+typology\phantom{000}  \\\hline
monolingual          & 4.36          & 4.31          &  \textbf{3.83} \phantom{0}($\downarrow$ 12.2\%) \\
+similar             & 5.73          & 4.93          & \textbf{4.24} \phantom{0}($\downarrow$ 26.0\%) \\
+dissimilar          & 5.88          & 4.98          & \textbf{4.41} \phantom{0}($\downarrow$ 25.0\%) \\
\end{tabular}
\caption{Perplexity experiments with \textsc{it} as test language.
Training languages: monolingual: \textsc{it}; +similar: \textsc{it, fr, ro}; +dissimilar: \textsc{it, fr, ro, hi}.
}
\label{tab:ppl-it}
\end{table}}
\begin{table}[ht]
\centering
\begin{tabular}{lccr}
&       \multicolumn{3}{c}{Perplexity ($\downarrow$)}     \\
\multicolumn{1}{c}{training set}  & baseline      & +lang         &+typology\phantom{000}  \\\hline
monolingual          & 4.36          & \multicolumn{1}{c}{--}         &  \multicolumn{1}{c}{--} \\
+similar             & 5.73          & 4.93          & \textbf{4.24} \phantom{0}($\downarrow$ 26.0\%) \\
+dissimilar          & 5.88          & 4.98          & \textbf{4.41} \phantom{0}($\downarrow$ 25.0\%) \\
\end{tabular}
\caption{Perplexity experiments with \textsc{it} as test language.
Training languages: monolingual: \textsc{it}; +similar: \textsc{it, fr, ro}; +dissimilar: \textsc{it, fr, ro, hi}.
}
\label{tab:ppl-it}
\end{table}

\ignore{\begin{table}[ht]
\centering
\begin{tabular}{lccr}
&       \multicolumn{3}{c}{Perplexity ($\downarrow$)}     \\
\multicolumn{1}{c}{} & baseline & +lang         & +typology\phantom{000}  \\\hline
monolingual          & 3.70     & 3.72          & \textbf{3.38} \phantom{0}($\downarrow$ \phantom{1}8.6\%) \\
+similar             & 4.14     & 3.78          & \textbf{3.35} \phantom{0}($\downarrow$ 19.1\%) \\
+dissimilar          & 4.29     & 3.82          & \textbf{3.42} \phantom{0}($\downarrow$ 20.3\%)
\end{tabular}
\caption{Perplexity experiments with \textsc{hi} as test language.
Training languages: monolingual: \textsc{hi}; +similar: \textsc{hi, ta, te}; +dissimilar: \textsc{hi, ta, te, en}.
}
\label{tab:ppl-hi}
\end{table}}

\begin{table}[ht]
\centering
\begin{tabular}{lccr}
&       \multicolumn{3}{c}{Perplexity ($\downarrow$)}     \\
\multicolumn{1}{c}{training set} & baseline & +lang         & +typology\phantom{000}  \\\hline
monolingual          & 3.70              & \multicolumn{1}{c}{--}         &  \multicolumn{1}{c}{--} \\
+similar             & 4.14     & 3.78          & \textbf{3.35} \phantom{0}($\downarrow$ 19.1\%) \\
+dissimilar          & 4.29     & 3.82          & \textbf{3.42} \phantom{0}($\downarrow$ 20.3\%)
\end{tabular}
\caption{Perplexity experiments with \textsc{hi} as test language.
Training languages: monolingual: \textsc{hi}; +similar: \textsc{hi, ta, te}; +dissimilar: \textsc{hi, ta, te, en}.
}
\label{tab:ppl-hi}
\end{table}

We see several patterns of results. First, polyglot models require, unsurprisingly, information about what language they are predicting to obtain good modeling performance. Second, typological information is more valuable than letting the model learn representations of the language along with the characters. 
Finally, typology-augmented polyglot models outperform
their monolingual baseline, providing evidence in support of the hypothesis that cross-lingual evidence is useful not only for learning cross-lingual representations and models, but monolingual ones as well.

\subsection{Lexical borrowing experiments}
\label{borrowing:experiments}
We fully reproduced lexical borrowing models described in \cite{borrowing-jair}
for three language pairs: \textsc{ar--sw}, \textsc{fr--ro}, and \textsc{it--mt}.
Train and test corpora are donor--loanword pairs in the language pairs.
Corpora statistics are given in \tref{tab:borrowing-stats} (note that these are extremely small data sets; thus small numbers of highly informative features a necessary for good generalization).
We use the reproduced systems as the baselines, and compare these
to the corresponding systems augmented with phone vectors, as described in \Sref{borrowing}.
\begin{table}[ht]
\centering
\begin{tabular}{lrrr}
\multicolumn{1}{c}{}  & \textsc{ar--sw}     & \textsc{fr--ro}       & \textsc{it--mt}   \\\hline
train	& 417  & 282 & 425 \\
test	& 73  & 50 & 75 \\
\end{tabular}
\caption{Number of training and test pairs the the borrowing datasets.}
\label{tab:borrowing-stats}
\end{table}

Integrated vectors were obtained from a single polyglot model with typology, trained on all languages with IPA dictionaries. 
For comparison with the results in \tref{tab:ppl-it}, perplexity of the model on the
\textsc{it} dataset (used for evaluation is \Sref{ppl}) is 4.16, even lower
than in the model trained on four languages.
To retrain the high-level conceptual linguistic features learned by the borrowing models,
we initialized the augmented systems with feature weights learned by the baselines,
and retrained. Final weights were established using cross-validation.
Then, we evaluated the accuracy of the augmented borrowing systems on the
held-out test data.

Accuracies are shown in \tref{tab:borrowing}.
We observe improvements of up to 5\% in accuracies of
\textsc{fr--ro}  and \textsc{it--mt} pairs.
Effectiveness of the same polyglot model trained on multiple
languages and integrated in different downstream systems
supports our assumption that the model remains stable and effective with addition of languages.
Our model is less effective for the \textsc{ar--sw} language pair.
We speculate that the results are worse, because
this is a pair of (typologically) more distant languages;
consequently, the phonological adaptation processes that happen in loanword
assimilation are more complex than mere substitutions of similar phones
that we are targeting via the integration of phone vectors.
\begin{table}[ht]
\centering
\begin{tabular}{lccr}
&       \multicolumn{3}{c}{Accuracy ($\uparrow$)}     \\
\multicolumn{1}{c}{}  & \textsc{ar--sw}     & \textsc{fr--ro}       & \textsc{it--mt}   \\\hline
baseline			& \textbf{48.4} & 75.6 & 83.3 \\
+multilingual	&  46.9 & \textbf{80.6} & \textbf{87.1} \\
\end{tabular}
\caption{Accuracies of the baseline models of lexical borrowing and the models augmented with phone vectors.
In all the experiments, we use vectors from a single Polyglot LM model trained on \textsc{ar, sw, fr, ro, it, mt}.}
\label{tab:borrowing}
\end{table}


\subsection{Speech synthesis experiments}
\label{voice:experiments}

\ignore{
We created pronunciation dictionaries for Hindi, Tamil, Telugu and English using the UniTran front end. The UniTran front end exploits a universal transliteration resource \cite{qian2010python} that transliterates Unicode code points into phonemes from the XSAMPA phone set. Our implementation of the UniTran front end assigns a single phoneme to a Unicode character, which can be inaccurate if the language has ambiguities in letter-to-sound conversion. In case of the Indian languages, this front end does not take care of phenomena such as schwa deletion in Hindi, voicing ambiguities in Tamil etc. In case of English, it assigns a single phoneme to letters with ambiguities in pronunciation, such as 'c'. We used a list of word types from the TTS databases for Hindi, Tamil and Telugu from the IIIT-H voice databases \cite{Prahallad2012} and the RMS US English voice \cite{Kominek2004} databases to create the pronunciation dictionaries by looking up the letter-to-sound rules in UniTran front end.}

\ignore{
TTS systems are evaluated using a variety of objective and subjective metrics. Subjective metrics, which require humans to rate or compare systems by listening to them can be expensive and time consuming \mfar{can be removed}.} 
A popular objective metric for measuring the quality of synthetic speech is the Mel Cepstral Distortion (MCD) \cite{hu2008evaluation}. The MCD metric calculates an L2 norm of the Mel Frequency Cepstral Coefficients (MFCCs) of natural speech from a held out test set, and synthetic speech generated from the same test set. Since this is a distance metric, a lower value of MCD suggests better synthesis. The MCD is a database-specific metric, but experiments by Kominek et al. \cite{kominek2008synthesizer} have shown that a decrease in MCD of 0.08 is perceptually significant, and a decrease of 0.12 is equivalent to doubling the size of the TTS database. In our experiments, we use MCD to measure the relative improvement obtained by our techniques.

We conducted experiments on the IIIT-H Hindi voice database \cite{Prahallad2012}, a 2 hour single speaker database recorded by a professional male speaker. We used the same front end (UniTran) to build all the Hindi TTS systems, with the only difference between the systems being the presence or absence of phonetic features and our vectors. For all our voice-based experiments, we built CLUSTERGEN Statistical Parametric Synthesis voices \cite{black2006clustergen} using the Festvox voice building tools \cite{black2003building} and the Festival speech synthesis engine \cite{Black1997}. 

The baseline TTS system was built using no phonetic features. We also built a TTS system with standard hand-crafted phonetic features. 
Table \ref{tab:voice-unitran} shows the MCD for the \textsc{hi} baseline,
the standard TTS with hand-crafted features, and augmented TTS systems built using monolingual and multilingual phone vectors constructed with Polyglot LMs.

\ignore{
\begin{table}[ht]
\centering
\begin{tabular}{lc}
&       \multicolumn{1}{c}{\textsc{mcd} ($\downarrow$)} \\
baseline 		& 4.27 \\\hline
+monolingual	& 4.15 \\
+multilingual	& 4.16 \\\hline
+hand-crafted			& \textbf{4.13} \\
\end{tabular}
\caption{MCD for voices built with the Indic front end}
\label{tab:voice-indic}
\end{table}

We see that in case of the voices and vectors built with the Indic front end, we get an improvement of 0.12 in MCD when we use monolingual vectors, and a slightly worse MCD when we use multilingual vectors. Using hand crafted phonetic features leads to the highest improvement in MCD of 0.14.}
\begin{table}[ht]
\centering
\begin{tabular}{lc}
&       \multicolumn{1}{c}{\textsc{mcd} ($\downarrow$)} \\
baseline 		& 4.58 \\\hline
+monolingual	& 4.40 \\
+multilingual	& \textbf{4.39} \\\hline
+hand-crafted			& 4.41 \\
\end{tabular}
\caption{MCD for the \textsc{hi} TTS systems.
Polyglot LM training languages: monolingual: \textsc{hi}; +multilingual: \textsc{hi, ta, te, en}.}
\label{tab:voice-unitran}
\end{table}

Our multilingual vectors outperform the baseline, with a significant decrease of 0.19 in MCD. Crucially, TTS systems augmented with the Polyglot LM phone vectors outperform also the standard TTS with hand-crafted features. We found that using both feature sets added no value, suggesting that learned phone vectors are capturing information that is equivalent to the hand-engineered vectors.

\subsection{Qualitative analysis of vectors}
\label{qvec:experiments}
Phone vectors learned by Polyglot LMs are mere sequences of real numbers.
An interesting question is whether these vectors capture linguistic (phonological)
qualities of phones they are encoding.
To analyze to what extent our vectors capture linguistic properties
of phones, we use the \textsc{qvec}---a tool to quantify and interpret linguistic 
content of vector space models \cite{qvec:enmlp:15}.
The tool aligns dimensions in a matrix of learned distributed representations with
dimensions of a hand-crafted linguistic matrix.
Alignments are induced via correlating columns in the distributed and the linguistic
matrices. To analyze the content of the distributed matrix, annotations from the
linguistic matrix are projected via the maximally-correlated alignments.

We constructed a phonological matrix in which
5,059 rows are IPA phones and 21 columns are boolean indicators of universal
phonological properties, e.g. \textit{consonant}, \textit{voiced},
\textit{labial}.\footnote{This matrix is described in \newcite{Littell-et-al:2016} and is available at \url{https://github.com/dmort27/panphon/}.}
We the projected annotations from the linguistic matrix and manually
examined aligned dimensions in the phone vectors from \Sref{borrowing:experiments}
(trained on six languages).
In the maximally-correlated columns---corresponding to linguistic features
\textit{long}, \textit{consonant}, \textit{nasalized}---we examined phones
with highest coefficients. These were: [\textipa{5:, U:, i:, O:, E:}] for \textit{long};
[\textipa{v, \textltailn , \t{dZ},  d,  f,  j, \t{ts}, N}] for \textit{consonant};
and [\textipa{\~O, \~E, \~A, \~ \oe}] for \textit{nasalized}.
 Clearly, the learned representation discover standard phonological features.
 Moreover, these top-ranked sounds are not grouped by a single language, e.g.,
 /\textipa{\t{dZ}}/ is present in Arabic but not in French, and
 /\textipa{\textltailn, N}/ are present in French but not in Arabic.
From this analysis, we conclude that (1) the model discovers linguistically meaningful phonetic features;
(2) the model induces meaningful related groupings across languages.

\section{Related Work}
\label{related}
\paragraph{Multilingual language models.}
Interpolation of monolingual LMs is an alternative to obtain a
multilingual model \cite{harbeck1997multilingual,weng1997study}.
However, interpolated models still require
a trained model per language, and do not allow parameter sharing at training time.
Bilingual language models trained on concatenated corpora were
explored mainly in speech recognition \cite{ward1998towards,wang2002towards,fugen2003efficient}.
Adaptations have been proposed to apply language models in bilingual settings
in machine translation \cite{niehues2011wider} and code switching \cite{adel2013codeswitching}.
These approaches, however, require adaptation to every pair of languages,
and an adapted model cannot be applied to more than two languages.

Independently, \newcite{ammar:16b} used a different 
polyglot architecture for multilingual dependency parsing. 
This work has also confirmed the utility of polyglot architectures 
in leveraging multilinguality.

\paragraph{Multimodal neural language models.}
Multimodal language modeling is integrating image/video modalities in text LMs.
Our work is inspired by the neural
multimodal LMs \cite{kiros2013multimodal,kiros2015multimodal}, which defined language models conditional on visual contexts, although we use a different language model architecture (recurrent vs. log-bilinear) and a different approach to gating modality.

\ignore{
\paragraph{Acoustic and phonetic features in applications.}
Recent work for improving prosody of a DNN-based TTS system supplements standard frame-level features extracted from text with features extracted from the acoustics at the sentence level \cite{watts2015sentence}. There has also been some work on using context embeddings produced by a DNN-based system for selecting models in an HMM-based synthesizer \cite{merritt2015deep}. }

\ignore{
\paragraph{Character-based NLMs.}
character language models find semantic representation
adapted to account for syntax \cite{ling2015finding,ballesteros2015improved},
morphology \cite{botha2014compositional,luong2013morphology,kim2015character}
but not phonology }

\section{Conclusion}
We presented a novel \textit{multilingual} language model architecture.
The model obtains substantial gains in perplexity, and improves
downstream text and speech applications.
Although we focus on phonology, our approach is general, and can be applied
in problems that integrate divergent modalities,
e.g., topic modeling, and multilingual tagging and parsing.


\section*{Acknowledgments}
This work was supported by the National Science Foundation through award IIS-1526745 and in part by the Defense Advanced Research Projects Agency (DARPA) Information Innovation Office (I2O). Program: Low Resource Languages for Emergent Incidents (LORELEI). Issued by DARPA/I2O under Contract No. HR0011-15-C-0114.

\bibliography{naaclhlt2016}

\begin{thebibliography}{}

\bibitem[\protect\citename{Adel \bgroup et al.\egroup
  }2013]{adel2013codeswitching}
Heike Adel, Ngoc~Thang Vu, and Tanja Schultz.
\newblock 2013.
\newblock Combination of recurrent neural networks and factored language models
  for code-switching language modeling.
\newblock In {\em Proc. ACL}, pages 206--211.

\bibitem[\protect\citename{Ammar \bgroup et al.\egroup }2016]{ammar:16b}
Waleed Ammar, George Mulcaire, Miguel Ballesteros, Chris Dyer, and Noah~A.
  Smith.
\newblock 2016.
\newblock Many languages, one parser.
\newblock {\em CoRR}, abs/1602.01595.

\bibitem[\protect\citename{Bansal \bgroup et al.\egroup }2014]{bansal:2014}
Mohit Bansal, Kevin Gimpel, and Karen Livescu.
\newblock 2014.
\newblock Tailoring continuous word representations for dependency parsing.
\newblock In {\em Proc. ACL}.

\bibitem[\protect\citename{Bengio \bgroup et al.\egroup
  }1994]{bengio1994learning}
Yoshua Bengio, Patrice Simard, and Paolo Frasconi.
\newblock 1994.
\newblock Learning long-term dependencies with gradient descent is difficult.
\newblock {\em IEEE Transactions on Neural Networks}, 5(2):157--166.

\bibitem[\protect\citename{Black and Lenzo}2003]{black2003building}
Alan~W Black and Kevin~A Lenzo.
\newblock 2003.
\newblock Building synthetic voices.
\newblock \url{http://festvox.org/bsv/}.

\bibitem[\protect\citename{Black and Taylor}1997]{Black1997}
Alan~W Black and Paul Taylor.
\newblock 1997.
\newblock The {F}estival speech synthesis system: system documentation.
\newblock Technical report, Human Communication Research Centre, University of
  Edinburgh.

\bibitem[\protect\citename{Black}2006]{black2006clustergen}
Alan~W Black.
\newblock 2006.
\newblock {CLUSTERGEN}: a statistical parametric synthesizer using trajectory
  modeling.
\newblock In {\em Proc. Interspeech}.

\bibitem[\protect\citename{Dryer and Haspelmath}2013]{wals2013}
Matthew~S. Dryer and Martin Haspelmath, editors.
\newblock 2013.
\newblock {\em WALS Online}.
\newblock Max Planck Institute for Evolutionary Anthropology.
\newblock \url{http://wals.info/}.

\bibitem[\protect\citename{Dyer \bgroup et al.\egroup
  }2015]{dyer2015transition}
Chris Dyer, Miguel Ballesteros, Wang Ling, Austin Matthews, and Noah~A Smith.
\newblock 2015.
\newblock Transition-based dependency parsing with stack long short-term
  memory.
\newblock In {\em Proc. ACL}.

\bibitem[\protect\citename{Faruqui and Dyer}2014]{faruqui:2014}
Manaal Faruqui and Chris Dyer.
\newblock 2014.
\newblock Improving vector space word representations using multilingual
  correlation.
\newblock In {\em Proc. EACL}.

\bibitem[\protect\citename{F{\"u}gen \bgroup et al.\egroup
  }2003]{fugen2003efficient}
Christian F{\"u}gen, Sebastian Stuker, Hagen Soltau, Florian Metze, and Tanja
  Schultz.
\newblock 2003.
\newblock Efficient handling of multilingual language models.
\newblock In {\em Proc. ASRU}, pages 441--446.

\bibitem[\protect\citename{Graves}2013]{graves:2013}
Alex Graves.
\newblock 2013.
\newblock Generating sequences with recurrent neural networks.
\newblock {\em CoRR}, abs/1308.0850.

\bibitem[\protect\citename{Guo \bgroup et al.\egroup }2014]{guo2014revisiting}
Jiang Guo, Wanxiang Che, Haifeng Wang, and Ting Liu.
\newblock 2014.
\newblock Revisiting embedding features for simple semi-supervised learning.
\newblock In {\em Proc. EMNLP}.

\bibitem[\protect\citename{Harbeck \bgroup et al.\egroup
  }1997]{harbeck1997multilingual}
Stefan Harbeck, Elmar N{\"o}th, and Heinrich Niemann.
\newblock 1997.
\newblock Multilingual speech recognition.
\newblock In {\em Proc. 2nd SQEL Workshop on Multi-Lingual Information
  Retrieval Dialogs}, pages 9--15.

\bibitem[\protect\citename{Haspelmath}2009]{haspelmath2009lexical}
Martin Haspelmath.
\newblock 2009.
\newblock Lexical borrowing: concepts and issues.
\newblock {\em Loanwords in the World's Languages: a comparative handbook},
  pages 35--54.

\bibitem[\protect\citename{Hermann and Blunsom}2014]{Hermann:2014:ACLphil}
Karl~Moritz Hermann and Phil Blunsom.
\newblock 2014.
\newblock {Multilingual Models for Compositional Distributional Semantics}.
\newblock In {\em Proc. ACL}.

\bibitem[\protect\citename{Hochreiter and Schmidhuber}1997]{hochreiter:1997}
Sepp Hochreiter and J\"urgen Schmidhuber.
\newblock 1997.
\newblock Long short-term memory.
\newblock {\em Neural Computation}, 9(8):1735--1780.

\bibitem[\protect\citename{Hu and Loizou}2008]{hu2008evaluation}
Yi~Hu and Philipos~C Loizou.
\newblock 2008.
\newblock Evaluation of objective quality measures for speech enhancement.
\newblock {\em Audio, Speech, \& Language Processing}, 16(1):229--238.

\bibitem[\protect\citename{Huang \bgroup et al.\egroup
  }2015]{huang-EtAl:2015:EMNLP}
Kejun Huang, Matt Gardner, Evangelos Papalexakis, Christos Faloutsos, Nikos
  Sidiropoulos, Tom Mitchell, Partha~P. Talukdar, and Xiao Fu.
\newblock 2015.
\newblock Translation invariant word embeddings.
\newblock In {\em Proc. EMNLP}, pages 1084--1088.

\bibitem[\protect\citename{Kingma and Ba}2014]{kingma2014adam}
Diederik Kingma and Jimmy Ba.
\newblock 2014.
\newblock Adam: A method for stochastic optimization.
\newblock {\em CoRR}, abs/1412.6980.

\bibitem[\protect\citename{Kiros and Salakhutdinov}2013]{kiros2013multimodal}
Ryan Kiros and Ruslan Salakhutdinov.
\newblock 2013.
\newblock Multimodal neural language models.
\newblock In {\em Proc. NIPS Deep Learning Workshop}.

\bibitem[\protect\citename{Kiros \bgroup et al.\egroup
  }2015]{kiros2015multimodal}
Ryan Kiros, Ruslan Salakhutdinov, and Richard Zemel.
\newblock 2015.
\newblock Unifying visual-semantic embeddings with multimodal neural language
  models.
\newblock {\em TACL}.

\bibitem[\protect\citename{Klakow and Peters}2002]{klakow2002testing}
Dietrich Klakow and Jochen Peters.
\newblock 2002.
\newblock Testing the correlation of word error rate and perplexity.
\newblock {\em Speech Communication}, 38(1):19--28.

\bibitem[\protect\citename{Kominek \bgroup et al.\egroup
  }2008]{kominek2008synthesizer}
John Kominek, Tanja Schultz, and Alan~W Black.
\newblock 2008.
\newblock Synthesizer voice quality of new languages calibrated with mean {M}el
  {C}epstral {D}istortion.
\newblock In {\em Proc. SLTU}, pages 63--68.

\bibitem[\protect\citename{Lazaridou \bgroup et al.\egroup
  }2013]{lazaridou:2013}
Angeliki Lazaridou, Eva~Maria Vecchi, and Marco Baroni.
\newblock 2013.
\newblock Fish transporters and miracle homes: How compositional distributional
  semantics can help {NP} parsing.
\newblock In {\em Proc. EMNLP}.

\bibitem[\protect\citename{Lewis \bgroup et al.\egroup }2015]{ethnologue2015}
M~Paul Lewis, Gary~F Simons, and Charles~D Fennig.
\newblock 2015.
\newblock {\em Ethnologue: Languages of the world}.
\newblock Texas: SIL International.
\newblock \url{http://www.ethnologue.com}.

\bibitem[\protect\citename{Ling \bgroup et al.\egroup }2015]{ling2015finding}
Wang Ling, Tiago Lu{\'\i}s, Lu{\'\i}s Marujo, Ram{\'o}n~Fernandez Astudillo,
  Silvio Amir, Chris Dyer, Alan~W Black, and Isabel Trancoso.
\newblock 2015.
\newblock Finding function in form: Compositional character models for open
  vocabulary word representation.
\newblock In {\em Proc. NAACL}.

\bibitem[\protect\citename{Littell \bgroup et al.\egroup
  }2016]{Littell-et-al:2016}
Patrick Littell, David Mortensen, Kartik Goyal, Chris Dyer, and Lori Levin.
\newblock 2016.
\newblock Bridge-language capitalization inference in {Western} {Iranian}:
  {Sorani}, {Kurmanji}, {Zazaki}, and {Tajik}.
\newblock In {\em Proceedings of the Eleventh International Conference on
  Language Resources and Evaluation (LREC'16)}.

\bibitem[\protect\citename{Lu \bgroup et al.\egroup }2015]{lu:2015}
Ang Lu, Weiran Wang, Mohit Bansal, Kevin Gimpel, and Karen Livescu.
\newblock 2015.
\newblock Deep multilingual correlation for improved word embeddings.
\newblock In {\em Proc. NAACL}.

\bibitem[\protect\citename{Mikolov \bgroup et al.\egroup
  }2010]{mikolov2010rnnlm}
Tomas Mikolov, Martin Karafi{\'a}t, Lukas Burget, Jan Cernock{\`y}, and Sanjeev
  Khudanpur.
\newblock 2010.
\newblock Recurrent neural network based language model.
\newblock In {\em Proc. Interspeech}, pages 1045--1048.

\bibitem[\protect\citename{Moran \bgroup et al.\egroup }2014]{phoible2014}
Steven Moran, Daniel McCloy, and Richard Wright, editors.
\newblock 2014.
\newblock {\em PHOIBLE Online}.
\newblock Max Planck Institute for Evolutionary Anthropology.
\newblock \url{http://phoible.org/}.

\bibitem[\protect\citename{Niehues \bgroup et al.\egroup
  }2011]{niehues2011wider}
Jan Niehues, Teresa Herrmann, Stephan Vogel, and Alex Waibel.
\newblock 2011.
\newblock Wider context by using bilingual language models in machine
  translation.
\newblock In {\em Proc. WMT}, pages 198--206.

\bibitem[\protect\citename{Prahallad \bgroup et al.\egroup
  }2012]{Prahallad2012}
Kishore Prahallad, E.~Naresh Kumar, Venkatesh Keri, S.~Rajendran, and Alan~W
  Black.
\newblock 2012.
\newblock The {IIIT-H} {I}ndic speech databases.
\newblock In {\em Proc. Interspeech}.

\bibitem[\protect\citename{Qian \bgroup et al.\egroup }2010]{qian2010python}
Ting Qian, Kristy Hollingshead, Su-youn Yoon, Kyoung-young Kim, Richard Sproat,
  and Malta LREC.
\newblock 2010.
\newblock A {P}ython toolkit for universal transliteration.
\newblock In {\em Proc. LREC}.

\bibitem[\protect\citename{Socher \bgroup et al.\egroup }2013]{socher:2013}
Richard Socher, Alex Perelygin, Jean Wu, Jason Chuang, Christopher~D. Manning,
  Andrew~Y. Ng, and Christopher Potts.
\newblock 2013.
\newblock Recursive deep models for semantic compositionality over a sentiment
  treebank.
\newblock In {\em Proc. EMNLP}.

\bibitem[\protect\citename{Sundermeyer \bgroup et al.\egroup
  }2012]{sundermeyer2012lstm}
Martin Sundermeyer, Ralf Schl{\"u}ter, and Hermann Ney.
\newblock 2012.
\newblock {LSTM} neural networks for language modeling.
\newblock In {\em Proc. Interspeech}.

\bibitem[\protect\citename{Thomason and Kaufman}2001]{thomason2001language}
Sarah~Grey Thomason and Terrence Kaufman.
\newblock 2001.
\newblock {\em Language contact}.
\newblock Edinburgh University Press Edinburgh.

\bibitem[\protect\citename{Tsvetkov and Dyer}2015]{borrowing-acl}
Yulia Tsvetkov and Chris Dyer.
\newblock 2015.
\newblock Lexicon stratification for translating out-of-vocabulary words.
\newblock In {\em Proc. ACL}, pages 125--131.

\bibitem[\protect\citename{Tsvetkov and Dyer}2016]{borrowing-jair}
Yulia Tsvetkov and Chris Dyer.
\newblock 2016.
\newblock Cross-lingual bridges with models of lexical borrowing.
\newblock {\em JAIR}, 55:63--93.

\bibitem[\protect\citename{Tsvetkov \bgroup et al.\egroup }2015]{qvec:enmlp:15}
Yulia Tsvetkov, Manaal Faruqui, Wang Ling, Guillaume Lample, and Chris Dyer.
\newblock 2015.
\newblock Evaluation of word vector representations by subspace alignment.
\newblock In {\em Proc. EMNLP}.
\newblock \url{https://github.com/ytsvetko/qvec}.

\bibitem[\protect\citename{Turian \bgroup et al.\egroup }2010]{turian:2010}
Joseph Turian, Lev Ratinov, and Yoshua Bengio.
\newblock 2010.
\newblock Word representations: a simple and general method for semi-supervised
  learning.
\newblock In {\em Proc. ACL}.

\bibitem[\protect\citename{Wang \bgroup et al.\egroup }2002]{wang2002towards}
Zhirong Wang, Umut Topkara, Tanja Schultz, and Alex Waibel.
\newblock 2002.
\newblock Towards universal speech recognition.
\newblock In {\em Proc. ICMI}, page 247.

\bibitem[\protect\citename{Wang \bgroup et al.\egroup
  }2015]{wang2015predicting}
Xin Wang, Yuanchao Liu, Chengjie Sun, Baoxun Wang, and Xiaolong Wang.
\newblock 2015.
\newblock Predicting polarities of tweets by composing word embeddings with
  long short-term memory.
\newblock In {\em Proc. ACL}, pages 1343--1353.

\bibitem[\protect\citename{Ward \bgroup et al.\egroup }1998]{ward1998towards}
Todd Ward, Salim Roukos, Chalapathy Neti, Jerome Gros, Mark Epstein, and Satya
  Dharanipragada.
\newblock 1998.
\newblock Towards speech understanding across multiple languages.
\newblock In {\em Proc. ICSLP}.

\bibitem[\protect\citename{Watanabe and Sumita}2015]{watanabe2015transition}
Taro Watanabe and Eiichiro Sumita.
\newblock 2015.
\newblock Transition-based neural constituent parsing.
\newblock In {\em Proc. ACL}.

\bibitem[\protect\citename{Watts \bgroup et al.\egroup
  }2015]{watts2015sentence}
Oliver Watts, Zhizheng Wu, and Simon King.
\newblock 2015.
\newblock Sentence-level control vectors for deep neural network speech
  synthesis.
\newblock In {\em Proc. Interspeech}.

\bibitem[\protect\citename{Weng \bgroup et al.\egroup }1997]{weng1997study}
Fuliang Weng, Harry Bratt, Leonardo Neumeyer, and Andreas Stolcke.
\newblock 1997.
\newblock A study of multilingual speech recognition.
\newblock In {\em Proc. EUROSPEECH}, pages 359--362.

\end{thebibliography}
\bibliographystyle{naaclhlt2016}

\end{document}